\documentclass{article}
\usepackage{ijcai25}

\usepackage{times}
\usepackage{soul}
\usepackage{url}
\usepackage[hidelinks]{hyperref}
\usepackage[utf8]{inputenc}
\usepackage[small]{caption}
\usepackage{graphicx}
\usepackage{amsmath}
\usepackage{amsthm}
\usepackage{booktabs}
\usepackage{algorithm}
\usepackage{algpseudocode}
\usepackage[switch]{lineno}
\usepackage{tcolorbox}
\usepackage{multirow}
\usepackage{mathrsfs}
\usepackage{subfig}
\usepackage{hyperref} 
\usepackage{amssymb}
\usepackage{bm}
\usepackage{newfloat}
\usepackage{listings}
\usepackage{amssymb}
\usepackage{algpseudocode}
\newtheorem{Assumption}{Assumption}
\newtheorem{Definition}{Definition}

\usepackage[top=2cm, bottom=2cm, left=2cm, right=2cm]{geometry}
\allowdisplaybreaks[4]
\floatname{listing}{Listing}

\title{Unveiling and Causalizing CoT: A Causal Pespective}

\author{
Jiarun Fu$^1$
\and
Lizhong Ding$^1$\and
Hao Li$^1$\and
Pengqi Li$^{1}$\and
Qiuning Wei$^1$\and
Xu Chen $^2$
\affiliations
$^1$Beijing Institute of Technology\\
$^2$Renmin University of China\\
\emails
3120235197@bit.edu.cn
}

\begin{document}

\maketitle

\begin{abstract}

Although Chain-of-Thought (CoT) has achieved remarkable success in enhancing the reasoning ability of large language models (LLMs), the mechanism of CoT remains a ``black box''. Even if the correct answers can frequently be obtained, existing CoTs struggle to make the reasoning understandable to human. In this paper, we unveil and causalize CoT from a causal perspective to ensure both correctness and understandability of all reasoning steps (to the best of our knowledge, the first such). We model causality of CoT via structural causal models (SCM) to unveil the reasoning mechanism of CoT. To measure the causality of CoT, we define the CoT Average Causal Effect (CACE) to test the causal relations between steps. For those steps without causality (wrong or unintelligible steps), we design a role-playing causal query algorithm to causalize these steps, resulting  a causalized CoT with all steps correct and understandable. Experimental results on both open-source and closed-source LLMs demonstrate that the causal errors commonly in steps are effectively corrected and the reasoning ability of LLMs is significantly improved. 

\end{abstract}

\section{Introduction}

\begin{figure*}
    \centering
    \includegraphics[width=0.8\linewidth]{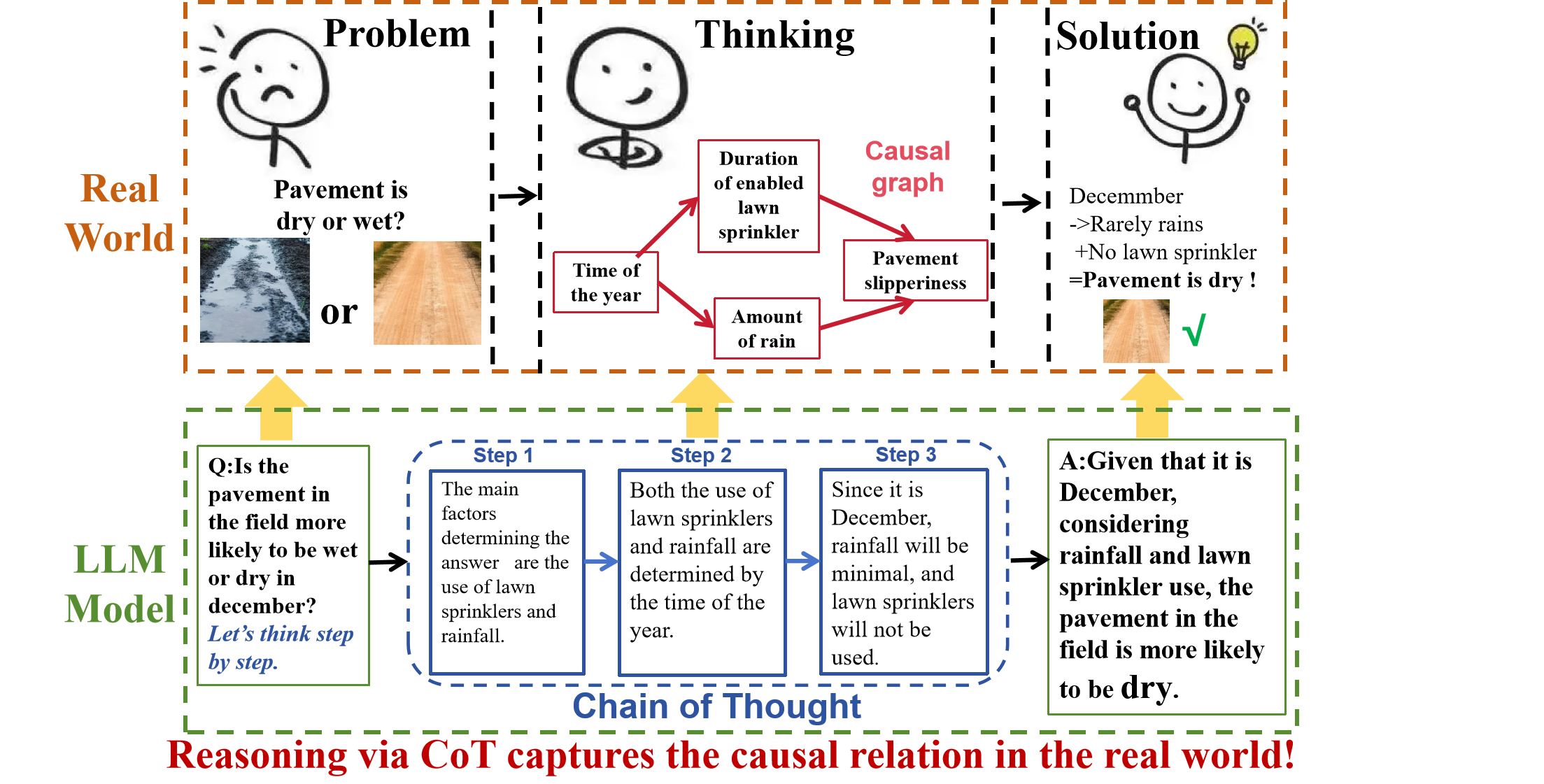}
    \caption{We assume the ability of CoT to reason correctly stems from its reflection of real-world causal relationships. As shown in the figure above, in the real world, based on our life experience, we infer the causal graph between variables such as month, rainfall, sprinkler use, and pavement slipperiness, then use this graph to deduce the answer. Similarly, LLMs employ Chain-of-Thought (CoT) to perform a reasoning process that aligns with these real-world causal relationships, ultimately arriving at the correct answer.}
    \label{CW}
\end{figure*}

\begin{quote}
``We do not have knowledge of a thing until we grasped its cause.''

\raggedleft---  Aristotle
\end{quote}

Causality \cite{pearl2009causality,yao2021survey} provides a unique perspective for exploring the mechanisms of machine learning algorithms, making significant contributions in areas such as trustworthy learning \cite{li2024debiased,xie2024generalized} and stable learning \cite{liu2023causal,wu2023stable,zhang2024causaldistillationalleviatingperformance}, etc. Large language models (LLMs) have made remarkable progresses in recent years \cite{wu2023brief,yang2024qwen2,liu2021gpt}, and it has been confirmed that LLMs can perform step-by-step reasoning through a Chain-of-Thought (CoT), which breaks down complex problems into a sequence of step-by-step thinking processes \cite{kojima2022large,hu2023tree,chen2024unlockingcapabilitiesthoughtreasoning}. Although CoT has led to remarkable achievements, it does not always provide positive outcomes \cite{sprague2024cot} and sometimes hinders reasoning performance \cite{lanham2023measuring}. So it is crucial to explore the mechanism of CoT and make it understandable. Therefore, analyzing CoT or LLMs' reasoning through causality offers a natural solution. \cite{chen2022disco,kiciman2023causal,bhattacharjee2024towards,jin2023can}.

However current research suggests LLMs behave like ``causal parrots'', merely reciting causal knowledge without truly understanding causality\cite{zevcevic2023causal,wu2024causality}, which limits the applicability of existing causal-based CoT methods, only in knowledge based tasks \cite{wu2024causality} and causal inference tasks \cite{jin2023cladder,zhang2024causal}. Beyond the community of causality, researches on the unveiling mechanism of CoT focus on the upper bound of reasoning ability \cite{feng2024towards,chen2024unlockingcapabilitiesthoughtreasoning} and contextual demonstration ability \cite{madaan2023what}, even so CoT remains a black box. We summarize the shortcomings of current researches as follows:

\begin{enumerate}
\item The lack of models for unveiling the mechanism of CoT to make the reasoning of LLMs interpretable.

\item The lack of algorithms that causalize all steps of CoT to make them correct and understandable.
\end{enumerate}

In this paper, we unveil the reasoning mechanism of CoT from a causal perspective and causalize CoT to make LLMs reasoning both correct and  understandable. Since causality is the most understandable and learnable logical relationship for humans \cite{rubin1980randomization,pearl2009causality,kaddour2022causal}, we assume that CoT's effectiveness in reasoning derives from its reflection of real-world causal relationships involved in problem-solving. The example in Figure ~\ref{CW} serves as a demonstration of this assumption. We construct a constructing structural causal model (SCM) of CoT to model the causal relations between reasoning steps, and further establish the SCM for unveiling mechanism of reasoning. In order to quantify the causality of CoT, we propose the CoT Average Causal Effect (CACE) to measure the causal relationship between steps of CoT from both answer and logic aspects. Based on empircial evidence, the first step of CoT largely determines the logic, we introduce the First-Step Causal Effect (FSCE) to quantify the casual logic of CoT reasoning. To test whether CoT has been causalized, we complete the causal inference of steps by CACE and FSCE to test the causality of each steps. For those CoTs have not been causalized, we apply a role-playing causal query algorithm including the causalizing process and the refine process to establish the causal logic for all steps. Through extensive experiments on both open-source and closed-source LLMs, we correct multiple causal error types between steps and improve significantly in capabilities of LLMs reasoning. 

Our main contributions are as follows:
\begin{enumerate}
    \item To the best of our knowledge, we are the first to unveils the mechanism of CoT through causality. We apply SCM for modeling the causality of CoT to make the reasoning of LLMs interpretable.
    \item In order to measure the causality of CoT, we propose Average Causal Effect (CACE) and First-Step Causal Effect (FSCE) to infer the causal logic between each reasoning step. Through causal inference, we summarize common causality errors in CoT.
    \item We propose a role-playing causal query algorithm to causalize reasoning steps that lack causal logic, ensuring the correctness and understandability of CoT.

\end{enumerate}

\section{Background}
Since SCM have been proven effective in exploring potential mechanisms, we introduce them to model the causality of CoT.

\subsection{Structural causal model (SCM)}
A structural causal model \cite{pearl2009causality,yao2021survey,kaddour2022causal} $\mathcal{M}$ is a 3-tuple $\langle V,U,\mathcal{F}\rangle$, where:

1. $U = \{u_1, ..., u_m\}$ is the set of exogenous variables, also called background variables, which are determined by factors outside the model;

2. $V = \{v_1, ..., v_n\}$ is the set of endogenous variables, which are determined by the variables in the model;

3. $\mathcal{F}=\{f_{v_1},...,f_{v_n}\}$ is the set of structural functions determining $V$, where $f_{v_i}:(pa(v_i),U)\mapsto v_i$ and $pa(v_i) \subseteq (V \setminus\{v_i\}) $. 

SCM are often presented for the case of intervention, which is denoted using $do$-$operator$  $do(\cdot)$ to implement $do(T=t)$ \cite{singh2020kernel,kaddour2022causal}, where T represens treatment and $t$ represents actual treatment content. Treatment effect is used to measure the impact of an interventional treatment, that is, the causal relationship between $t$ and the outcome of interest $Y$. The most widely used average treatment effect (ATE) is defined as $\gamma(t):=\mathbb{E}[Y \mid do(T=t)]$
and Conditional Average Treatment Effect (CATE) is defined as $\gamma(t,u_i):=\mathbb{E}[Y \mid do(T=t),u_i]$, where $ u_i \in U $ are exogenous variables. It is important to note that conditioning on $T = t$ means that we are looking at the outcomes for the subset of the population who have received the treatment $t$ in the observation data, so $\mathbb{E}\left[Y\mid do(T=t)\right]\neq \mathbb{E}\left[Y\mid T=t\right]$.

\subsection{Chain-of-Thought reasoning}
Referring to previous CoT work \cite{qiao2022reasoning,chu2024navigate,xiang2025towards}, we define the notations as follows: the question $Q$, the instruction $IS$, LLM $p_{\mathrm{LM}}$ and the answer $A=[a_1,a_2,\ldots,a_n]$. The model takes the $Q$ and $IS$ as inputs to produce the answer $A$ as its output:
\begin{equation}
\begin{aligned}
p ( A \mid IS, Q )=\prod_{i=1}^{n} p_{\mathrm{LM}} ( IS, Q, a_{<i} ) .
\end{aligned}
\end{equation}

LLM generates its reasoning paths $C=[c_1,c_2,...,c_n]$ by explaining them step by step before producing the final answer $A$:
\begin{equation*}
\begin{aligned}
\label{2}
p (A, C\mid IS, Q) = p (A \mid C,IS, Q) \cdot p(C\mid IS, Q),
\end{aligned}
\end{equation*}
\begin{equation*}
\begin{aligned}
p(C\mid IS, Q)=\prod_{i=1}^{n} p_{\mathrm{LM}} (  IS, Q, c_{<i} ),
\end{aligned}
\end{equation*}
\begin{equation*}
\begin{aligned}
p ( A \mid IS, Q, C)=\prod_{j=1}^{n} p_{\mathrm{LM}} ( IS, Q, C, a_{<j} ) .
\end{aligned}
\end{equation*}

\begin{figure}[H]
        \centering
        \includegraphics[width=1\linewidth]{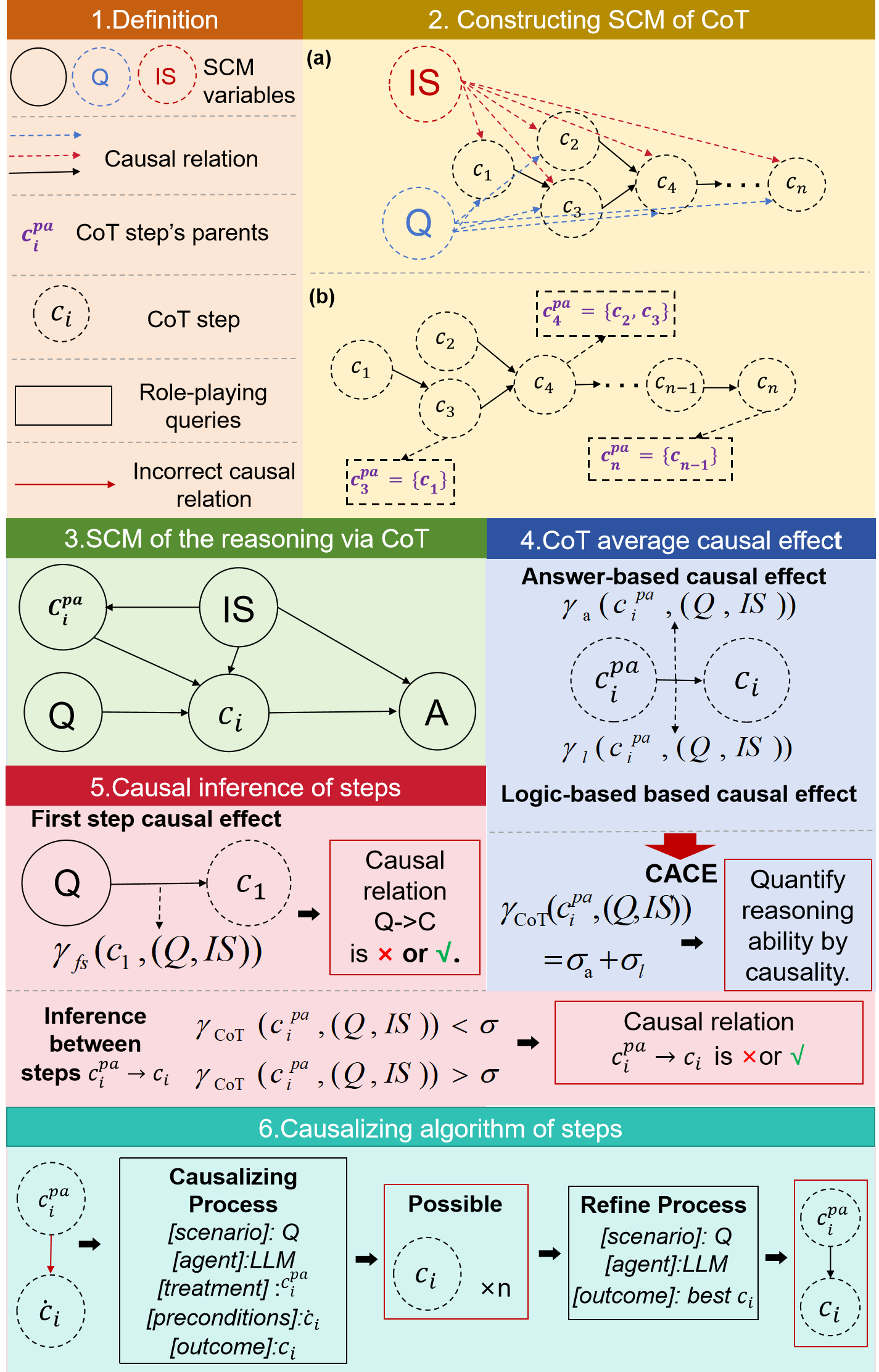}
        \caption{From modeling CoT to causalizing CoT.}
        \label{SCM}
\end{figure}\section{Opening the black box of CoT: By SCM modeling}

Since causality is one of the most intuitive forms of logic for humans, we aim to unveil the mechanism of CoT from a causal pespective. Building on this idea, we assume that the reasoning ability of CoT arises from real-world causality, where causal relationships exist between each step, thus reflecting the correct causal graph inherent in the problem's solution. We model causality of CoT via SCM to make LLMs' reasoning understandable. With reasonable assumptions on the causal structure, we define the CACE to measure the causality of CoT.

\subsection{Constructing SCM of CoT}
We treat $IS$ and $Q$ as exogenous variables in SCM since they do not change during reasoning. Then we define $C=[c_1,c_2,...,c_n]$ as endogenous variables, $p_{\mathrm{LM}}(\cdot)$ as structural functions. As we assume the CoT reflects the causal graph, $c_i$ should be influenced by its parent nodes $c^{pa}_i$, $Q$, and $IS$, where $c^{pa}_i$ is a subset of the power set $ 2^{[{c}_1,{c}_2,...,{c}_{i-1}]}$. Then we can get the constructing SCM of CoT as:

\begin{equation}
\begin{aligned}
c_i = p_{\mathrm{LM}}\left( IS,Q,{c}^{pa}_i\right).
\end{aligned}
\end{equation}

As shown in Figure~\ref{SCM}, we explain the detials of constructing SCM of CoT and the definition of $C^{pa}=[c_1^{pa},c_2^{pa},...,c_n^{pa}]$.

\subsection{SCM of the reasoning via CoT}
Based on the constructing SCM of CoT, we redefine the reasoning via CoT  fomular \eqref{2} to causal version to model CoT reasoning mechanism. We define LLMs answer the $Q$ with $a_i$ based on $IS$ and step-by-step reasoning path $c_i$ , which is generated based on $c_i^{pa}$. We format SCM of the reasoning via CoT as:

\begin{equation*}
\begin{aligned}
&p (a_i, c_i\mid IS, Q,c_i^{pa}) =
\\\
&p (a_i \mid c_i, IS, Q,c_i^{pa}) \cdot p(c_i\mid  IS, Q,c_i^{pa}).
\end{aligned}
\end{equation*}

Following the constructing SCM of CoT, we can define:
\begin{equation*}
\begin{aligned}
p(C\mid  IS, Q,C^{pa})=\prod_{i=1}^{n} p_{\mathrm{LM}} (IS, Q, c_{<i}^{pa}),
\end{aligned}
\end{equation*}
\begin{equation*}
\begin{aligned}
p ( A \mid IS, Q, C,C^{pa})=\prod_{j=1}^{n} p_{\mathrm{LM}}( IS, Q, C,C^{pa},a_{<j}).
\end{aligned}
\end{equation*}

We illustrate SCM of the reasoning via CoT in Figure~\ref{SCM}. One important point to note is that, according to the Markov rule of SCM, $c_i$ has a direct causal relationship only with its parent nodes ${c}^{pa}_i$, not with the previous steps $c{<i}$ from a traditional perspective. Therefore, the causal path $Q \rightarrow c_i \rightarrow A$ shows that $c_i$ can serve as a mediator between the question $Q$ and the answer $A$. Similarly, the causal relationship between $a_{i-1}$ and $a_i$ should only be reflected by $c_i$, otherwise, a collider bias will occur between $A$ and the parent nodes of $C$, ${C}^{pa}$.

\subsection{CoT average causal
effect}

Since we have successfully established a SCM for CoT, the subsequent question is how to quantify the causality within CoT. Since the causality of CoT should possess both logical coherence and correctness, we quantify the causal relationship $c_i^{pa} \rightarrow c_i$ based on the logic of reasoning and the correctness of answers corresponding to $a_i$. Using the $do(\cdot)$ operator, we define the logic-based average causal effect $\gamma_{l}$ and the answer-based average causal effect $\gamma_{a}$ as follows:

\begin{equation}
\label{9}
\begin{aligned}
&\gamma_{a}(c^{pa}_i,(Q, IS)) \\
=&\mathbb{E}\left[(a_i\mid IS,Q,do(c^{pa}_i))-(a_i\mid IS,Q,c^{pa}_i)\right],
\\
&\gamma_{l}(c^{pa}_i,(Q, IS)) \\
=&\mathbb{E}[({c}_i\mid IS,Q,do(c^{pa}_i))-({c}_i\mid IS,Q,c^{pa}_i)].
\end{aligned}
\end{equation}

Based on $\gamma_{l}$ and $\gamma_{a}$, we propose CoT average causal effect of CoT (CACE) $\gamma_\textrm{CoT}$ to quantify causal relationships of $c_i^{pa}\rightarrow c_i$:

\begin{equation}
\label{11}
\begin{aligned}
&\gamma_\textrm{CoT}(c^{pa}_i,(Q, IS)) \\
=&\gamma_{a}(c^{pa}_i,(Q, IS))+ \gamma_{l}(c^{pa}_i,(Q, IS)).
\end{aligned}
\end{equation}

\subsubsection{Assumptions for SCM of CoT}
To eliminate the well-known selection bias, we introduce three standard assumptions that are commonly relied upon in existing SCM \cite{rubin1980randomization,pearl2009causality,yao2021survey,litwo,zeng2024survey,jin2023cladder,zhang2024causal,wu2024causality}.

\begin{Assumption}[Stable Unit Treatment Value (SUTVA)]
The distribution of the $a_i$ and $c_i$ with $p_{\mathrm{LM}}(\cdot)$ are assumed to be independent of the interventions $do(c_i^{pa})$.
\end{Assumption}

Since $do(c_i^{pa})$ intervenes in the step of reasoning and modifies $p_{\mathrm{LM}}(\cdot)$, the distributions of $A$ and $C$ remain unaffected by this intervention. SUTVA ensures that equation \eqref{11} accurately reflects the causal relationships $c_i^{pa}\rightarrow c_i$ \cite{qi2023proximal,wu2023stable,zhang2024causaldistillationalleviatingperformance}.

\begin{Assumption}[Unconfoundedness]
The distribution of $do(c_i^{pa})$ is independent of $a_i$ and $c_i$, given exogenous variables $IS$,$Q$.
\end{Assumption}

Our second assumption extends the unconfoundedness assumption to equation \eqref{11}, assuming that there are no unobserved variables that have a causal relationships with the intervention. The unconfoundedness assumption ensures that the treatment of $c_i^{pa}$ accurately reflects the causal relationship $c_i^{pa}\rightarrow c_i$ \cite{xie2020generalized,qi2023proximal,zuo2024interventional}.

\begin{Assumption}[Overlap]
Every unit should have a nonzero probability to receive either treatment status. Formally, $0 < p(do(c_i^{pa})\mid Q, IS) < 1$.
\end{Assumption}
The overlap assumption is fundamental in the field of causal inference, which ensures the validity and feasibility of $c_i^{pa}$ in \eqref{11} \cite{litwo,zeng2024survey,jin2023cladder}.

By using $\gamma_\textrm{CoT}$, we can successfully quantify the causal relationships between steps in CoT, enabling us to test whether reasoning steps have causal logic.

\section{CauCoT: Causalized Chain-of-Thought}
We have successfully constructed the SCM for CoT to demonstrate the causality within its reasoning mechanism. However, to make LLMs reasoning both correct and understandable, we still need a method to causalize CoT in order to ensure steps of CoT both correct and casually logical. In this section, we propose our method Causalized CoT (CauCoT), including causal inference of steps and causalizing algorithm of steps. 

\subsection{Causal inference of steps }
Since steps of causalized CoT are both correct and understandable, to test whether there is correct causal relation between steps, we need to quantify the causality inherent in CoT from both the correctness of answer and the logic of reasoning aspects. Based on \eqref{11}, we can effectively measure the causal relationships between the steps of CoT. However, we still lack a method to measure the causal logic between CoT and the question $Q$ (since $IS$ is assumed to have an equal causal effect on all SCM components except for $Q$, it does not require separate analysis). Based on empircial evidence ,if the first step, $c_1$, does not have causality to answer $Q$, subsequent steps will be unable to has causal logic since they are built on incorrect basic (this will be further discussed in the Appendix~\ref{A2}). Therefore, we propose the First-Step Causal Effect (FSCE), denoted as $\gamma_{fs}$, to measure the causality between CoT and the question. We define $\gamma_{fs}$ as:

\begin{equation}
\label{8}
\begin{aligned}
&\gamma_{{fs}}\left(c_1,(Q, IS)\right) \\
=&\mathbb{E}[(a_1\mid IS,Q,do(c_1))-(a_1\mid IS,Q,c_1)].
\end{aligned}
\end{equation}

To ensure uniformity of CauCoT, we apply LLMs to finish $\mathbb{E}(\cdot)$ by scoreing, prompt examples are as follow:

\begin{tcolorbox}[colback=white, colframe=red!75!black, title=The prompt of $\gamma_{fs}$]
Now that we try to answer the question $Q$ step by step and $c_1$ is the first step. Now you need to carefully evaluate the impact of $c_1$ to answer $Q$ correctly. Please the impact and the full score is 100. 
\end{tcolorbox}  

\begin{tcolorbox}[colback=white, colframe=red!75!black, title=The prompt of $\gamma_{a}$]
Now that we try to answer the question $Q$ step by step through reasoning path $c_i$. Now you need to carefully evaluate the impact of $c^{pa}_i$ to answer $Q$ based on $c_i$ is final step. Please the impact and the full score is 100. 
\end{tcolorbox}  

\begin{tcolorbox}[colback=white, colframe=red!75!black, title=The prompt of $\gamma_{l}$]
Now that we try to answer the question $Q$ step by step through reasoning path $c_i$. Now you need to carefully evaluate the impact of $c^{pa}_i$ to generate $c_i$. If the full score is 100, you need to score the size of the impact.
\end{tcolorbox}  

To complete the causal inference of steps , we still need a criterion for determining whether the causality exists. Therefore, we define $\sigma$ as the causalized confidence degree and provide the following definition:

\begin{Definition}
\label{def:CCFH}
We say there exists a causal relation between $c_i$ and $c^{pa}_i$ if
\begin{equation}
\gamma_\mathrm{CoT}(c^{pa}_i,(Q, IS))<\sigma.
\end{equation} 
\end{Definition}

To ensure the flexibility of causal inference of steps  we redefine \eqref{11} as :
\begin{equation}
\label{3}
\begin{aligned}
&\gamma_\textrm{CoT}(c^{pa}_i,(Q, IS)) \\
=&\alpha\gamma_{a}(c^{pa}_i,(Q, IS))+\beta \gamma_{l}(c^{pa}_i,(Q, IS)),
\end{aligned}
\end{equation}
where $\alpha$ and $\beta$ are hyperparameters (the setting of $\alpha$ and $\beta$ will be further discussed in the experiment.), with $\alpha + \beta = 1$. These parameters can be used to adjust the emphasis on different aspects of the causal relationship, such as prioritizing the reasoning logic or the correct answer. It is important to note that $\sigma$ is not a strictly fixed value; it can be adjusted based on the specific scenario. For example, in mathematical reasoning, where strict logic between steps is emphasized, a higher value for $\sigma$ should be used. In contrast, for knowledge reasoning, which may rely more on continuous causality between steps, a relatively lower value for $\sigma$ may be appropriate. In summary, $\sigma$ can be adjusted according to the type of task to align with the requirements of different reasoning tasks. The settings for $\sigma$ will be discussed further in Appendix~\ref{A2},~\ref{A4}.

\subsection{Causalizing algorithm of steps}
In the steps of causal inference, we have completed the causal inference in CoT and obtained a set of CoT including steps without the causality, wthih defined as $\sigma = [\dot{C}_1, \dot{C}_2, \dots, \dot{C}_n]$. Then we need to causalize those CoT to make every steps has causal relation. We define causal relation as follow:

\begin{Definition}
\label{CCausality}
We say $C$ has correct causality to answer $Q$ when any $c_i\in C$  has causal relation with $c^{pa}_i$, and $c_i$ is faithful to answer the qusetion $Q$.
\end{Definition}

Inspired by \cite{han2024mining}, where role-playing queries generate more targeted and unique responses, we apply a two-step algorithm consisting of role-playing causal queries, which includes the causalizing process and the refinement process, to establish the causal relationships in all steps. 

\subsubsection{Causalizing process}
The definition of the roles in the causalizing Process is as follows: LLMs act as the [agent], attempting to find the correct reasoning steps. The agent depends on the [scenario] $Q$, such as an economist in an economic $Q$ or a medical professional in a medical $Q$. $c^{pa}_i$ represents the [treatment], $[c_i]$ is the [outcome], and the current incorrect steps $\dot{c_i}\in \dot{C}_i$ is [preconditions]. Therefore, our role-playing causal queries ask the LLM to assume the role of the  [agent]and predict the [outcome] after the [treatment], given the [preconditions] in a hypothetical [scenario]. The [treatment] and [preconditions] provided assist the LLM in better capturing the correct causal relationships, as we are presenting incorrect answers based on the [preconditions]. An example of the prompt used in the causalizing process is as follows:
\begin{tcolorbox}[colback=white, colframe=blue!75!black, title=Causalizing Process]
You are [agent] who is answering question $Q$ and trying to think about the problem step by step, where $c^{pa}_i$ and $\dot{c_i}$ should have strong causal relation to answer the question correctly. Since $\dot{c_i}$ is wrong step, now generate $c_i$ that can meet the strong causal relationship with the previous step $c^{pa}_i$, make $c^{pa}_i\rightarrow c_i$ is correct reasoning path to correctly answer $Q$. Please list $[c_i]$ chains with strong credibility and explain your result.
\end{tcolorbox}  
\subsubsection{Refine process}
Applying refining, the large language model conducts self-reviews during the process of generating answers to verify the correctness and consistency of its responses. In the our refine process, we prompt the LLM to rethink its previous steps and select the most accurate and faithful response to answer the question $Q$, 
\begin{tcolorbox}[colback=white, colframe=blue!75!black, title=Refine process]
You are [agent] who is answering question $Q$, as previously described. Among the $[c_i]$ listed above, choose chain $c_i$ that are most likely to have strong reasoning to answering question $Q$. For the chosen chain, explain the reasoning.
\end{tcolorbox}  
\subsubsection{CauCoT algorithm: Loop until Causalized}
Although we introduce a refine process to enhance the quality and efficiency of causalizing algorithm of steps, making all steps of CoT causalized is still a challenging task. Therefore, we further propose the CauCoT algorithm to ensure the reliability of the causalizing of steps.

\begin{algorithm}
    \caption{CauCoT Algorithm}
    \label{alg:algorithm}
    \textbf{Input}: CoT $C$, prompt $IS$, Question $Q$, causalized confidence degree $\sigma$, and large language model $p_\mathrm{LM}$.\\
    \textbf{Output}: Causalized $C$
    \begin{algorithmic}[1]
        \State Initialize large language model $p_{\mathrm{LM}}$ with prompt $IS$.
        \For{each $c_i$ in $C$}
            \If{$i = 1$}
                \State Finish $\gamma_{{fs}}$ inference by \eqref{8}
                \While{$\gamma_{{fs}}\left(c_1, (Q, IS)\right) < \epsilon$}
                    \State Do causalizing algorithm of steps with $\dot{c_1}$
                \EndWhile
                \State \textbf{continue}
            \EndIf
            \State Finish $\gamma_\textrm{CoT}(c^{pa}_i, (Q, IS))$ by \eqref{11}
            \While{$\gamma_\textrm{CoT}(c^{pa}_i, (Q, IS)) < \sigma$}
                \State Do causalizing algorithm of steps with $\dot{c_i}$
            \EndWhile
        \EndFor
    \end{algorithmic}
\end{algorithm}

\begin{table*}
  \caption{EM results of reasoning QA. The first row of the table indicates the content of each column. The first column lists the datasets, the second column shows the LLMs used for testing (including different versions of Qwen, Deepseek, and Llama). Columns three to five display the EM results of the baseline models we compared, while the sixth column presents the results of our method, CauCoT. We highlight the top three results in \textcolor{red}{red} (first), \textcolor{blue}{blue} (second), and \textcolor{purple}{purple} (third).}
  \label{table1}
  \centering
  \begin{tabular}{llllll}
  \toprule
 \textbf{Dataset}& \textbf{Model}     & \textbf{0-shot}  &\textbf{CoT} & \textbf{PB} & \textbf{CauCoT}\\
    \midrule
  GSM8K& Qwen2.5-3b   & \textcolor{purple}{0.485} & \textcolor{blue}{0.791} & 0.415  &  \textcolor{red}{0.848}\\
  &Qwen2.5-7b  &  \textcolor{purple}{0.517} & \textcolor{blue}{0.854}  & 0.425  &  \textcolor{red}{0.871}\\
  & Llama3-8b    &   \textcolor{purple}{0.499}  & \textcolor{blue}{0.845} & 0.433   & \textcolor{red}{0.864}\\
  & Qwen2.5-72b  &  \textcolor{purple}{0.789}	& \textcolor{blue}{0.915} &  0.500  & \textcolor{red}{0.943}\\
  & Deepseek-v3-37B  &  \textcolor{purple}{0.850} & \textcolor{blue}{0.915} &  0.500  & \textcolor{red}{0.950}\\
  \midrule
  Math & Qwen2.5-3b   &   \textcolor{purple}{0.426}  & \textcolor{blue}{0.472} & 0.376  &  \textcolor{red}{0.638}\\
  & Qwen2.5-7b   &  \textcolor{purple}{0.498}  & \textcolor{blue}{0.576} & 0.448 & \textcolor{red}{0.677} \\
   & Llama3-8b    &   0.443   & \textcolor{blue}{0.519} & \textcolor{purple}{0.462}   & \textcolor{red}{0.653}\\
   & Qwen2.5-72b  &  \textcolor{purple}{0.570}  & \textcolor{blue}{0.820} & 0.499 & \textcolor{red}{0.882}     \\
   & Deepseek-v3-37B  &  \textcolor{purple}{0.616}   & \textcolor{blue}{0.902} & 0.500 & \textcolor{red}{0.935} \\
  \midrule
  Olympiadbench & Qwen2.5-3b   &  0.062 &  \textcolor{purple}{0.106} &  \textcolor{blue}{0.316} &  \textcolor{red}{0.388} \\
  & Qwen2.5-7b   &  0.076 &  \textcolor{purple}{0.153} & \textcolor{blue}{0.390} &  \textcolor{red}{0.512} \\
   & Llama3-8b    &   0.069   & \textcolor{purple}{0.133} & \textcolor{blue}{0.356}   & \textcolor{red}{0.442}\\
  & Qwen2.5-72b   & 0.115 & \textcolor{purple}{0.197} & \textcolor{blue}{0.499} & \textcolor{red}{0.634}  \\
  & Deepseek-v3-37B  &  0.179 & \textcolor{purple}{ 0.254} &  \textcolor{blue}{0.500}  & \textcolor{red}{0.665}\\
  \midrule
  Omnimath & Qwen2.5-3b   & 0.142 & \textcolor{purple}{0.165}  & \textcolor{blue}{0.350} &  \textcolor{red}{0.452}\\
   &  Qwen2.5-7b & 0.181  & \textcolor{purple}{0.242} &  \textcolor{blue}{0.388} &  \textcolor{red}{0.556} \\
   & Llama3-8b  & 0.161   & \textcolor{purple}{0.251} & \textcolor{blue}{ 0.352}   & \textcolor{red}{0.441}\\
   & Qwen2.5-72b   & 0.217 & 	\textcolor{purple}{0.362} & \textcolor{blue}{0.498} &  \textcolor{red}{0.675} \\
  & Deepseek-v3-37B  &  0.309 & \textcolor{purple}{0.381} &  \textcolor{blue}{0.500}  & \textcolor{red}{0.722}\\
    \bottomrule
  \end{tabular}
\end{table*}

\section{Experimental results and analyses}

We evaluate the causalization performance of CauCoT on the Polluted CoT dataset using both open-source and closed-source LLMs. First, we analyze four common causal errors in CoT that CauCoT addresses. Subsequently, through experiments about evaluation of reasoning and causality, we prove that CauCoT demonstrate that the causal errors commonly in steps are effectively corrected and the reasoning ability of LLMs is significantly improved.

\subsection{Datasets}

We select the PROCESSBENCH (PB) dataset \cite{processbench}. It consists of 3,400 tests cases, primarily focused on competition and Olympiad-level math problems. Each test case contains a step-by-step solution with error location annotated by human experts, and half of the data is causalized, while the other half contains causal errors.

Specifically, it contains sample questions from the following four datasets: \textbf{GSM8K} \cite{cobbe2021training},\textbf{Math} \cite{hendrycks2021measuring},\textbf{OlympiadBench} \cite{he2024olympiadbench} and \textbf{Omni-MATH} \cite{gao2024omni} (Details will be show in Appendix ~\ref{Data}).

\subsection{Common causal errors in CoT}

Through experiments, we verified that CauCoT can effectively correct causal errors in Polluted CoT. We summarize the potential causal errors in CoT as follows (The detailed analysis and example wii be showed in Appendix~\ref{Eerror}):

\textbf{Causality measure error \cite{chwialkowski2014wild,scheines2017measurement}.}

Causality measure error refers to the inaccuracy in causal inference caused by the incorrect measurement or transmission of causal variables within the CoT.

\textbf{Collider error \cite{schneider2020collider,holmberg2022collider}.}

Collider error is an error caused by the inability to accurately measure the impact of two variables CoT on answering the question (the next step is the collision point).

\textbf{Causal sensitivity error \cite{cinelli2019sensitivity}.}

Causal sensitivity error refers to the failure to accurately measure the impact of two causal variables in CoT on answering the question. It occurs when the next step fails to account for the sensitivity of the variables’ relationship, leading to an incorrect conclusion.

\textbf{Mediation error \cite{pearl2014interpretation}.}

Mediation error occurs when CoT fails to correctly identify the mediator variable, which is essential in explaining the causal relationship between other variables. This error arises when CoT ignores the role of the mediator, leading to a misunderstanding of establishing a direct causal relationship between two variables, resulting in an incorrect conclusion.

\subsection{Main results}
In this section, we evaluate the performance of CauCoT in reasoning questions. We analyze the experiments from two perspectives: the effectiveness of regular answers and the causality of the CoT. This dual analysis allows us to assess both the accuracy of the answers and the degree to which CauCoT ensures correct causal relationships in the reasoning process.

\subsubsection{Evaluations of reasoning ability.}
For baselines, we compare our method with the conventional CoT prompting method \cite{wei2022chain} and PROCESSBENCH (PB) \cite{processbench}, which help demonstrate the superiority of CauCoT and its effectiveness in explaining CoT's reasoning ability from a causal perspective. Additionally, a 0-shot QA method is used as a control group. We employ Exact Match (EM) as our evaluation metric, and all tasks are evaluated in a 0-shot setting.

\textbf{Comparison with Baselines.}

The reasoning QA results are shown in Table \eqref{table1}. As expected, CauCoT outperforms all other methods across all datasets and open-source large models. Notably, on more complex logical problem datasets, such as OlympiadBench and Omni-MATH, the improvement with CauCoT is more pronounced compared to relatively simpler datasets like GSM8K and Math. Another point to highlight is that, in datasets like GSM8K and Math, CoT, or even 0-shot, can outperform PB. However, when faced with more complex datasets like OlympiadBench and Omni-MATH, PB actually yields better performance. This is because PB contains a subset of data that has already been causalized, which aids LLMs in providing correct answers. In contrast, conventional CoT methods do not ensure causality, which can result in less than half of the answers being correct. CauCoT, however, shows significant improvement, confirming that causality is key to ensuring CoT can perform reasoning correctly.

\subsubsection{Evaluations of causality}

For the causality of CoT, we compare CauCoT with CoT and PB by analyzing the changes in causal effects between each step. Across four datasets, we report the heterogeneous effect (HE) and the factual average treatment effect (ATE). The heterogeneous effect is defined as: ($HE = \sqrt{n^{-1} \sum_{i=1}^{n} \left( \left( \hat{a}_{i}-a_{i}\mid do(c_i) \right) \right)^{2}}$). Additionally, we report the factual average treatment effect ($\hat{ATE}$) to evaluate the overall impact of the intervention on CoT reasoning.

\textbf{Comparison with Baselines.}

As shown in Figure 3, we analyze the causality between steps. CauCoT achieves the strongest causality across all datasets, and its inference EM is also the highest, confirming that CauCoT successfully causalizes CoT. When the causality between CoT steps is stronger than that in PB, its inference performance surpasses PB. Conversely, when the causality is weaker than that in PB, its inference performance lags behind CoT. It further reinforces the fact that causality is a critical determinant of CoT's reasoning ability.
\begin{figure}[H]
    \centering
    \includegraphics[width=1\linewidth]{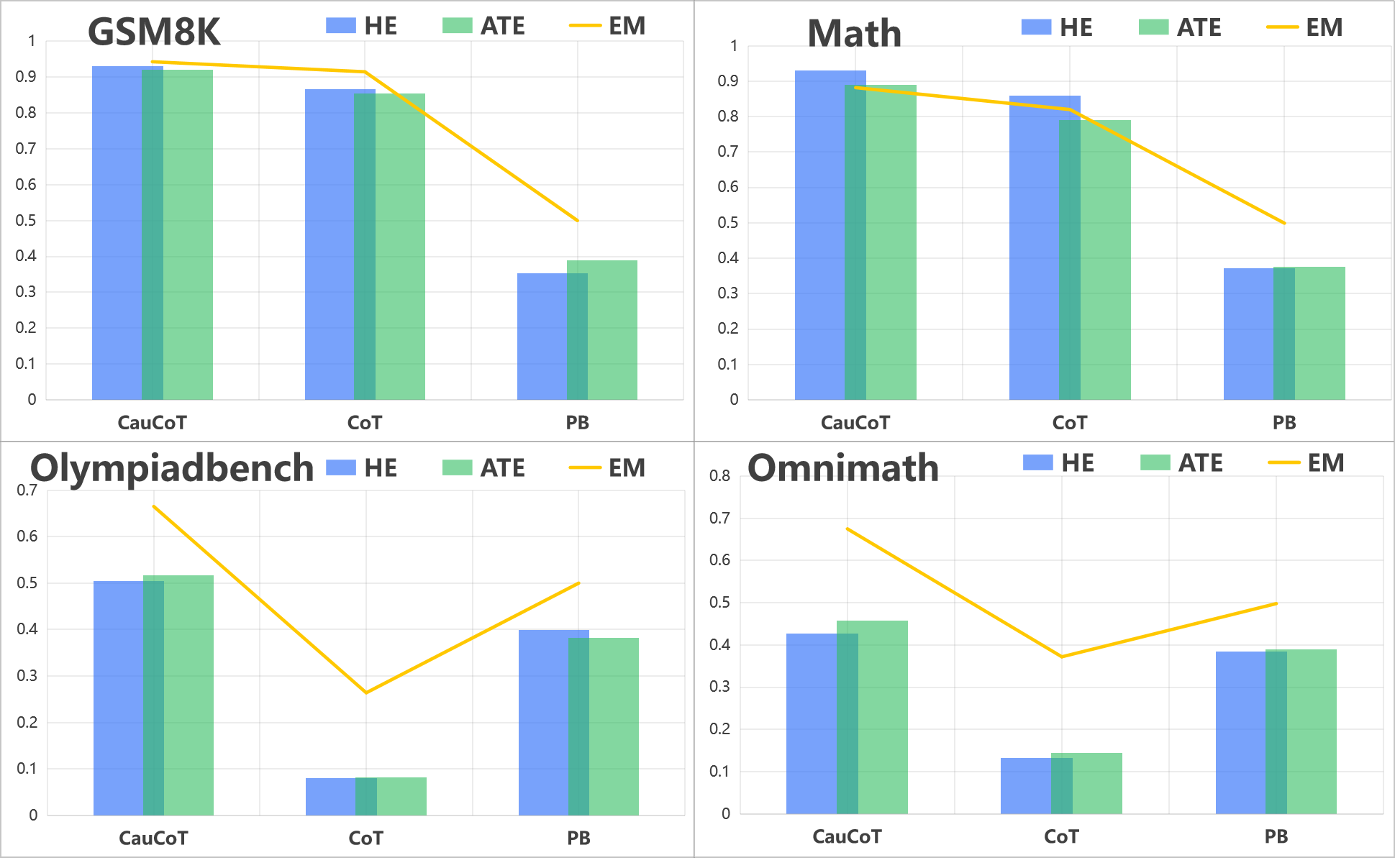}
    \caption{Causalized Evaluation on Qwen2.5-72B}
    \label{Qwen}
\end{figure}

\begin{figure}[H]
    \centering
    \includegraphics[width=1\linewidth]{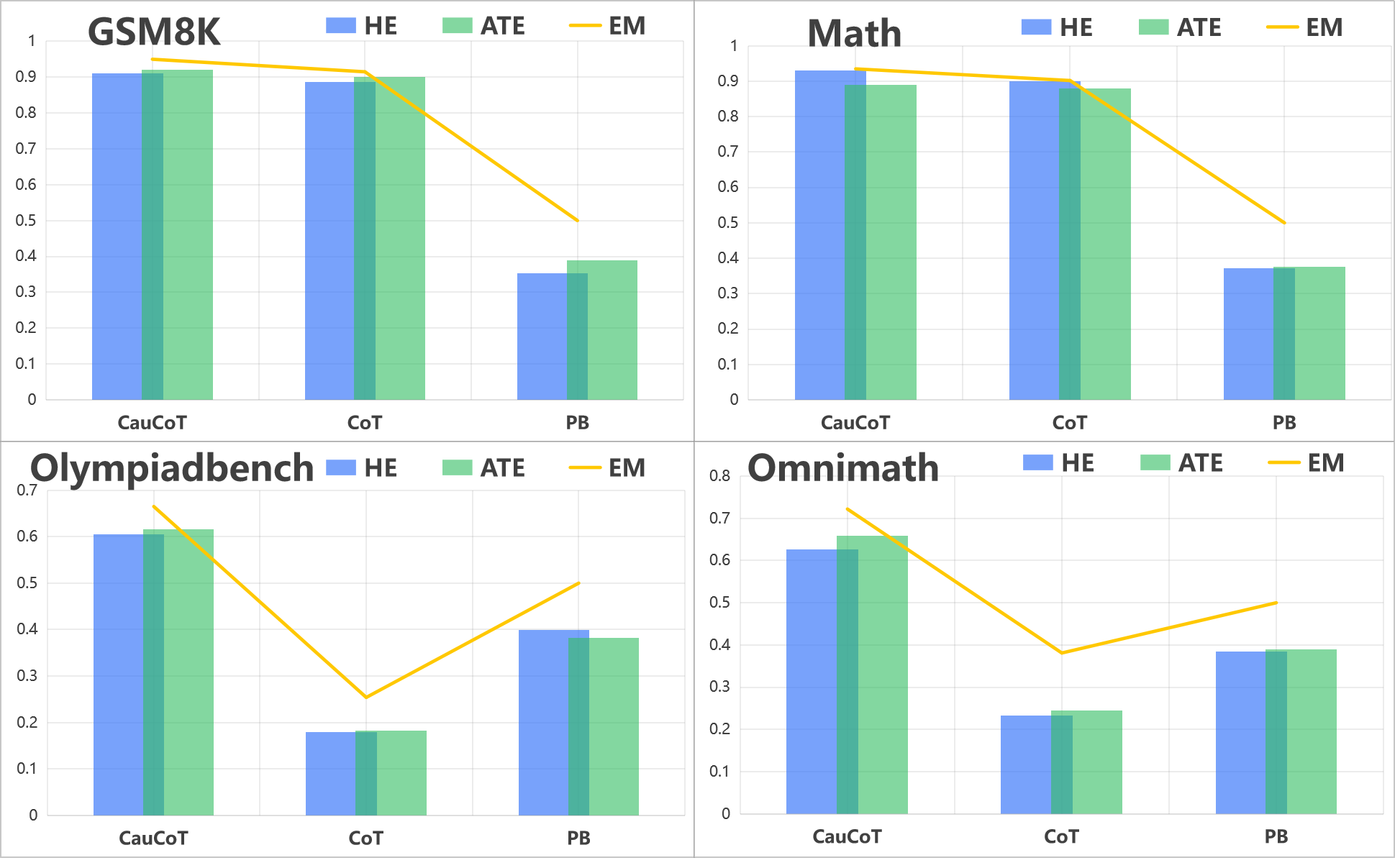}
    \caption{Causalized Evaluation on Deepseek-v3-37B}
    \label{Deepseek}
\end{figure}

\subsubsection{Hyperparamter experiments}
\label{A3}

\begin{figure}[H]
    \centering
    \includegraphics[width=1\linewidth]{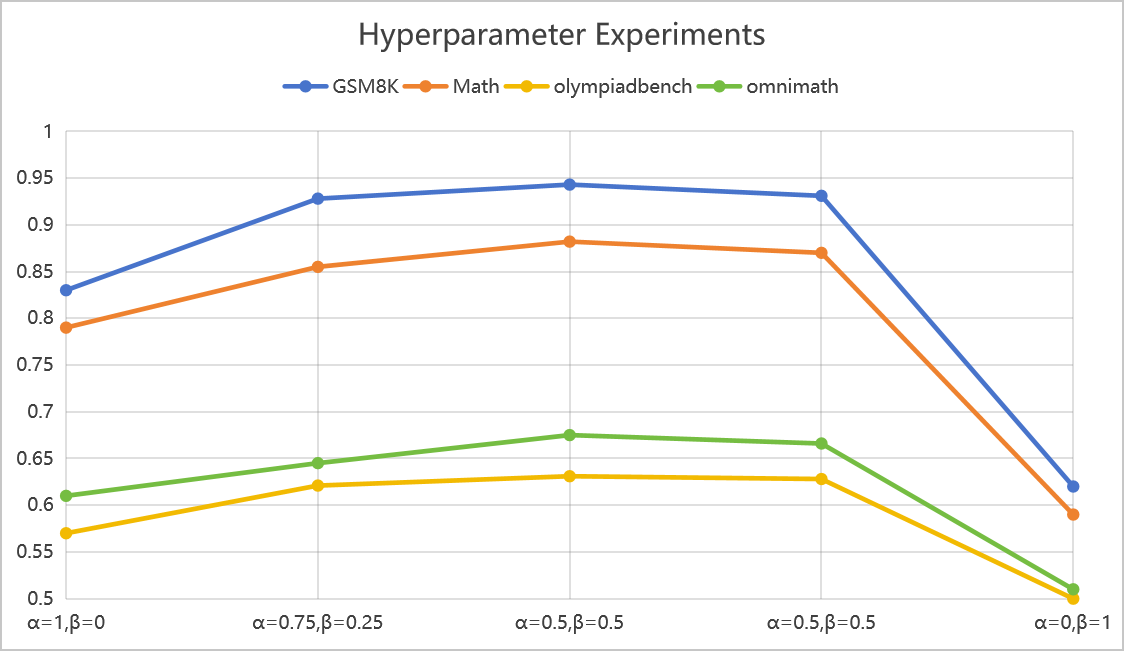}
    \caption{Hyperparamter experiments}
    \label{HP}
\end{figure}

Here, we discuss the setting of hyperparameters $\alpha$ and $\beta$ based on Qwen2.5-72b. As shown in the Figure~\ref{HP}, when both factors are balanced, CauCoT achieves the strongest causality. It is noteworthy that when $\alpha=0,\beta=1$, the performance drop of CauCoT is much larger than when $\alpha=1,\beta=0$. This indicates that in order to causalize CoT, causality should not only be considered from a logical perspective ($\gamma_l$), but the causal logic with answers ($\gamma_a$). This demonstrates that, to unveil the mechanisms of CoT and further enhance the performance of large models' reasoning, \textbf{correctness and understandability of steps are equally important.}

\subsection{Discussions of experimental results}
We make the following discussion based on experimental results:

1.Based on the evaluations of reasoning ability and causality, the stronger the causality between reasoning steps, the stronger the reasoning ability of CoT is. This proves that \textbf{our proposed SCM successfully unveils the mechanisms of CoT reasoning from a causal perspective, making the reasoning of LLMs interpretable.}

2.CauCoT successfully to correct the causal errors in the polluted dataset, and CauCoT achieves the strongest reasoning ability and causality across all datasets and LLMs. Thus, \textbf{we successfully achieves an algorithms that causalize all steps of CoT to make them correct and understandable. }

We will also provide further analysis of the experiment results in the Appendix.

\label{A4}
\section{Conclusion}
We unveil and causalize CoT from a causal perspective, ensuring LLMs' reasoning both correct and understandable. We model the causality of CoT via SCM, illustrating that steps' reasoning ability stems from causality. To test whether CoT has been causalized, we proposed causal inference of steps from both logic and answer aspects. we design a role-playing causal query algorithm to causalize steps without causality, resulting on our method CauCoT. Experimental results on both open-source and closed-source LLMs demonstrate that CauCoT effectively correct the causal error in steps and improve the reasoning ability of LLMs. In summary, our proposed SCM provides a new paradigm for modeling LLMs' reasoning from a causal perspective. CauCoT achieves an unprecedented causalizing algorithms of CoT, offering a causal solution for enabling correct and understandable LLMs' reasoning. Our work also presents the causality community with an innovative methodology for causally analyzing the reasoning mechanisms of CoT or LLMs.

\bibliographystyle{named}
\bibliography{sample-base}

\begin{thebibliography}{}

\bibitem[\protect\citeauthoryear{Bhattacharjee \bgroup \em et al.\egroup }{2024}]{bhattacharjee2024towards}
Amrita Bhattacharjee, Raha Moraffah, Joshua Garland, and Huan Liu.
\newblock Towards llm-guided causal explainability for black-box text classifiers.
\newblock In {\em AAAI}, 2024.

\bibitem[\protect\citeauthoryear{Burns and Wieth}{2004}]{burns2004collider}
Bruce~D Burns and Mareike Wieth.
\newblock The collider principle in causal reasoning: why the monty hall dilemma is so hard.
\newblock {\em Journal of Experimental Psychology: General}, 133(3):434, 2004.

\bibitem[\protect\citeauthoryear{Chen \bgroup \em et al.\egroup }{2022}]{chen2022disco}
Zeming Chen, Qiyue Gao, Antoine Bosselut, Ashish Sabharwal, and Kyle Richardson.
\newblock Disco: Distilling counterfactuals with large language models.
\newblock {\em arXiv:2212.10534}, 2022.

\bibitem[\protect\citeauthoryear{Chen \bgroup \em et al.\egroup }{2024}]{chen2024unlockingcapabilitiesthoughtreasoning}
Qiguang Chen, Libo Qin, Jiaqi Wang, Jinxuan Zhou, and Wanxiang Che.
\newblock Unlocking the capabilities of thought: A reasoning boundary framework to quantify and optimize chain-of-thought.
\newblock In {\em NeurIPS}, 2024.

\bibitem[\protect\citeauthoryear{Chu \bgroup \em et al.\egroup }{2024}]{chu2024navigate}
Zheng Chu, Jingchang Chen, Qianglong Chen, Weijiang Yu, Tao He, Haotian Wang, Weihua Peng, Ming Liu, Bing Qin, and Ting Liu.
\newblock Navigate through enigmatic labyrinth a survey of chain of thought reasoning: Advances, frontiers and future.
\newblock In {\em Proceedings of the 62nd Annual Meeting of the Association for Computational Linguistics (Volume 1: Long Papers)}, 2024.

\bibitem[\protect\citeauthoryear{Chwialkowski \bgroup \em et al.\egroup }{2014}]{chwialkowski2014wild}
Kacper~P Chwialkowski, Dino Sejdinovic, and Arthur Gretton.
\newblock A wild bootstrap for degenerate kernel tests.
\newblock {\em Advances in neural information processing systems}, 27, 2014.

\bibitem[\protect\citeauthoryear{Cinelli \bgroup \em et al.\egroup }{2019}]{cinelli2019sensitivity}
Carlos Cinelli, Daniel Kumor, Bryant Chen, Judea Pearl, and Elias Bareinboim.
\newblock Sensitivity analysis of linear structural causal models.
\newblock In {\em ICML}, 2019.

\bibitem[\protect\citeauthoryear{Cobbe \bgroup \em et al.\egroup }{2021}]{cobbe2021training}
Karl Cobbe, Vineet Kosaraju, Mohammad Bavarian, Mark Chen, Heewoo Jun, Lukasz Kaiser, Matthias Plappert, Jerry Tworek, Jacob Hilton, Reiichiro Nakano, et~al.
\newblock Training verifiers to solve math word problems.
\newblock {\em arXiv:2110.14168}, 2021.

\bibitem[\protect\citeauthoryear{D{\'\i}az and van~der Laan}{2013}]{diaz2013sensitivity}
Iv{\'a}n D{\'\i}az and Mark~J van~der Laan.
\newblock Sensitivity analysis for causal inference under unmeasured confounding and measurement error problems.
\newblock {\em The international journal of biostatistics}, 9(2):149--160, 2013.

\bibitem[\protect\citeauthoryear{Feng \bgroup \em et al.\egroup }{2024}]{feng2024towards}
Guhao Feng, Bohang Zhang, Yuntian Gu, Haotian Ye, Di~He, and Liwei Wang.
\newblock Towards revealing the mystery behind chain of thought: a theoretical perspective.
\newblock In {\em NeurIPS}, 2024.

\bibitem[\protect\citeauthoryear{Gao \bgroup \em et al.\egroup }{2024}]{gao2024omni}
Bofei Gao, Feifan Song, Zhe Yang, Zefan Cai, Yibo Miao, Qingxiu Dong, Lei Li, Chenghao Ma, Liang Chen, Runxin Xu, et~al.
\newblock Omni-math: A universal olympiad level mathematic benchmark for large language models.
\newblock {\em arXiv:2410.07985}, 2024.

\bibitem[\protect\citeauthoryear{Han}{2024}]{han2024mining}
Sukjin Han.
\newblock Mining causality: Ai-assisted search for instrumental variables.
\newblock {\em arXiv:2409.14202}, 2024.

\bibitem[\protect\citeauthoryear{He \bgroup \em et al.\egroup }{2024}]{he2024olympiadbench}
Chaoqun He, Renjie Luo, Yuzhuo Bai, Shengding Hu, Zhen~Leng Thai, Junhao Shen, Jinyi Hu, Xu~Han, Yujie Huang, Yuxiang Zhang, et~al.
\newblock Olympiadbench: A challenging benchmark for promoting agi with olympiad-level bilingual multimodal scientific problems.
\newblock {\em arXiv:2402.14008}, 2024.

\bibitem[\protect\citeauthoryear{Hendrycks \bgroup \em et al.\egroup }{2021}]{hendrycks2021measuring}
Dan Hendrycks, Collin Burns, Saurav Kadavath, Akul Arora, Steven Basart, Eric Tang, Dawn Song, and Jacob Steinhardt.
\newblock Measuring mathematical problem solving with the math dataset.
\newblock {\em arXiv:2103.03874}, 2021.

\bibitem[\protect\citeauthoryear{Holmberg and Andersen}{2022}]{holmberg2022collider}
Mathias~J Holmberg and Lars~W Andersen.
\newblock Collider bias.
\newblock {\em Jama}, 327(13):1282--1283, 2022.

\bibitem[\protect\citeauthoryear{Hu \bgroup \em et al.\egroup }{2023}]{hu2023tree}
Mengkang Hu, Yao Mu, Xinmiao Yu, Mingyu Ding, Shiguang Wu, Wenqi Shao, Qiguang Chen, Bin Wang, Yu~Qiao, and Ping Luo.
\newblock Tree-planner: Efficient close-loop task planning with large language models.
\newblock {\em arXiv:2310.08582}, 2023.

\bibitem[\protect\citeauthoryear{Imai \bgroup \em et al.\egroup }{2010}]{imai2010identification}
Kosuke Imai, Luke Keele, and Teppei Yamamoto.
\newblock Identification, inference and sensitivity analysis for causal mediation effects.
\newblock {\em Statistical Science}, 25(1):51--71, 2010.

\bibitem[\protect\citeauthoryear{Jin \bgroup \em et al.\egroup }{2023a}]{jin2023cladder}
Zhijing Jin, Yuen Chen, Felix Leeb, Luigi Gresele, Ojasv Kamal, Zhiheng Lyu, Kevin Blin, Fernando Gonzalez, Max Kleiman-Weiner, Mrinmaya Sachan, et~al.
\newblock Cladder: assessing causal reasoning in language models.
\newblock In {\em NeurIPS}, 2023.

\bibitem[\protect\citeauthoryear{Jin \bgroup \em et al.\egroup }{2023b}]{jin2023can}
Zhijing Jin, Jiarui Liu, Zhiheng Lyu, Spencer Poff, Mrinmaya Sachan, Rada Mihalcea, Mona Diab, and Bernhard Sch{\"o}lkopf.
\newblock Can large language models infer causation from correlation?
\newblock {\em arXiv:2306.05836}, 2023.

\bibitem[\protect\citeauthoryear{Kaddour \bgroup \em et al.\egroup }{2022}]{kaddour2022causal}
Jean Kaddour, Aengus Lynch, Qi~Liu, Matt~J Kusner, and Ricardo Silva.
\newblock Causal machine learning: A survey and open problems.
\newblock {\em arXiv:2206.15475}, 2022.

\bibitem[\protect\citeauthoryear{Kojima \bgroup \em et al.\egroup }{2022}]{kojima2022large}
Takeshi Kojima, Shixiang~Shane Gu, Machel Reid, Yutaka Matsuo, and Yusuke Iwasawa.
\newblock Large language models are zero-shot reasoners.
\newblock {\em arXiv:2205.11916}, 2022.

\bibitem[\protect\citeauthoryear{Kıcıman \bgroup \em et al.\egroup }{2023}]{kiciman2023causal}
Emre Kıcıman, Robert Ness, Amit Sharma, and Chenhao Tan.
\newblock Causal reasoning and large language models: Opening a new frontier for causality.
\newblock {\em arXiv:2305.00050}, 2023.

\bibitem[\protect\citeauthoryear{Lanham \bgroup \em et al.\egroup }{2023}]{lanham2023measuring}
Tamera Lanham, Anna Chen, Ansh Radhakrishnan, Benoit Steiner, Carson Denison, Danny Hernandez, Dustin Li, Esin Durmus, Evan Hubinger, Jackson Kernion, et~al.
\newblock Measuring faithfulness in chain-of-thought reasoning.
\newblock {\em arXiv:2307.13702}, 2023.

\bibitem[\protect\citeauthoryear{Li \bgroup \em et al.\egroup }{2023}]{litwo}
Baohong Li, Anpeng Wu, Ruoxuan Xiong, and Kun Kuang.
\newblock Two-stage shadow inclusion estimation: An iv approach for causal inference under latent confounding and collider bias.
\newblock In {\em ICML}, 2023.

\bibitem[\protect\citeauthoryear{Li \bgroup \em et al.\egroup }{2024}]{li2024debiased}
Haoxuan Li, Chunyuan Zheng, Yanghao Xiao, Peng Wu, Zhi Geng, Xu~Chen, and Peng Cui.
\newblock Debiased collaborative filtering with kernel-based causal balancing.
\newblock {\em arXiv:2404.19596}, 2024.

\bibitem[\protect\citeauthoryear{Liu and Kuang}{2023}]{liu2023causal}
Chenxi Liu and Kun Kuang.
\newblock Causal structure learning for latent intervened non-stationary data.
\newblock In {\em ICML}, 2023.

\bibitem[\protect\citeauthoryear{Liu \bgroup \em et al.\egroup }{2021}]{liu2021gpt}
Xiao Liu, Yanan Zheng, Zhengxiao Du, Ming Ding, Yujie Qian, Zhilin Yang, and Jie Tang.
\newblock Gpt understands, too.
\newblock {\em arXiv:2103.10385}, 2021.

\bibitem[\protect\citeauthoryear{Madaan \bgroup \em et al.\egroup }{2023}]{madaan2023what}
Aman Madaan, Katherine Hermann, and Amir Yazdanbakhsh.
\newblock What makes chain-of-thought prompting effective? a counterfactual study.
\newblock In {\em The 2023 Conference on Empirical Methods in Natural Language Processing}, 2023.

\bibitem[\protect\citeauthoryear{Papana \bgroup \em et al.\egroup }{2011}]{Papana_2011}
A.~Papana, D.~Kugiumtzis, and P.~G. Larsson.
\newblock Reducing the bias of causality measures.
\newblock {\em Physical Review E}, 83(3), 2011.

\bibitem[\protect\citeauthoryear{Pearl}{2009}]{pearl2009causality}
Judea Pearl.
\newblock {\em Causality}.
\newblock Cambridge university press, 2009.

\bibitem[\protect\citeauthoryear{Pearl}{2012}]{pearl2012measurement}
Judea Pearl.
\newblock On measurement bias in causal inference.
\newblock {\em arXiv:1203.3504}, 2012.

\bibitem[\protect\citeauthoryear{Pearl}{2014}]{pearl2014interpretation}
Judea Pearl.
\newblock Interpretation and identification of causal mediation.
\newblock {\em Psychological methods}, 19(4), 2014.

\bibitem[\protect\citeauthoryear{Qi \bgroup \em et al.\egroup }{2023}]{qi2023proximal}
Zhengling Qi, Rui Miao, and Xiaoke Zhang.
\newblock Proximal learning for individualized treatment regimes under unmeasured confounding.
\newblock {\em Journal of the American Statistical Association}, 2023.

\bibitem[\protect\citeauthoryear{Qiao \bgroup \em et al.\egroup }{2022}]{qiao2022reasoning}
Shuofei Qiao, Yixin Ou, Ningyu Zhang, Xiang Chen, Yunzhi Yao, Shumin Deng, Chuanqi Tan, Fei Huang, and Huajun Chen.
\newblock Reasoning with language model prompting: A survey.
\newblock {\em arXiv:2212.09597}, 2022.

\bibitem[\protect\citeauthoryear{Rubin}{1980}]{rubin1980randomization}
Donald~B Rubin.
\newblock Randomization analysis of experimental data: The fisher randomization test comment.
\newblock {\em Journal of the American statistical association}, 1980.

\bibitem[\protect\citeauthoryear{Scheines and Ramsey}{2017}]{scheines2017measurement}
Richard Scheines and Joseph Ramsey.
\newblock Measurement error and causal discovery.
\newblock In {\em CEUR workshop proceedings}, 2017.

\bibitem[\protect\citeauthoryear{Schneider}{2020}]{schneider2020collider}
Eric~B Schneider.
\newblock Collider bias in economic history research.
\newblock {\em Explorations in Economic History}, 78, 2020.

\bibitem[\protect\citeauthoryear{Singh \bgroup \em et al.\egroup }{2020}]{singh2020kernel}
Rahul Singh, Liyuan Xu, and Arthur Gretton.
\newblock Kernel methods for causal functions: Dose, heterogeneous, and incremental response curves.
\newblock {\em arXiv:2010.04855}, 2020.

\bibitem[\protect\citeauthoryear{Sprague \bgroup \em et al.\egroup }{2024}]{sprague2024cot}
Zayne Sprague, Fangcong Yin, Juan~Diego Rodriguez, Dongwei Jiang, Manya Wadhwa, Prasann Singhal, Xinyu Zhao, Xi~Ye, Kyle Mahowald, and Greg Durrett.
\newblock To cot or not to cot? chain-of-thought helps mainly on math and symbolic reasoning.
\newblock {\em arXiv:2409.12183}, 2024.

\bibitem[\protect\citeauthoryear{Wei \bgroup \em et al.\egroup }{2022}]{wei2022chain}
Jason Wei, Xuezhi Wang, Dale Schuurmans, Maarten Bosma, Brian Ichter, Fei Xia, H~Chi, Quoc~V Le, and Denny Zhou.
\newblock Chain-of-thought prompting elicits reasoning in large language models.
\newblock {\em In NeurIPS}, 2022.

\bibitem[\protect\citeauthoryear{Wu \bgroup \em et al.\egroup }{2023a}]{wu2023stable}
Anpeng Wu, Kun Kuang, Ruoxuan Xiong, Bo~Li, and Fei Wu.
\newblock Stable estimation of heterogeneous treatment effects.
\newblock In {\em ICML}, 2023.

\bibitem[\protect\citeauthoryear{Wu \bgroup \em et al.\egroup }{2023b}]{wu2023brief}
Tianyu Wu, Shizhu He, Jingping Liu, Siqi Sun, Kang Liu, Qing-Long Han, and Yang Tang.
\newblock A brief overview of chatgpt: The history, status quo and potential future development.
\newblock {\em IEEE/CAA Journal of Automatica Sinica}, 10(5):1122--1136, 2023.

\bibitem[\protect\citeauthoryear{Wu \bgroup \em et al.\egroup }{2024}]{wu2024causality}
Anpeng Wu, Kun Kuang, Minqin Zhu, Yingrong Wang, Yujia Zheng, Kairong Han, Baohong Li, Guangyi Chen, Fei Wu, and Kun Zhang.
\newblock Causality for large language models.
\newblock {\em arXiv:2410.15319}, 2024.

\bibitem[\protect\citeauthoryear{Xiang \bgroup \em et al.\egroup }{2025}]{xiang2025towards}
Violet Xiang, Charlie Snell, Kanishk Gandhi, Alon Albalak, Anikait Singh, Chase Blagden, Duy Phung, Rafael Rafailov, Nathan Lile, Dakota Mahan, et~al.
\newblock Towards system 2 reasoning in llms: Learning how to think with meta chain-of-though.
\newblock {\em arXiv:2501.04682}, 2025.

\bibitem[\protect\citeauthoryear{Xie \bgroup \em et al.\egroup }{2020}]{xie2020generalized}
Feng Xie, Ruichu Cai, Biwei Huang, Clark Glymour, Zhifeng Hao, and Kun Zhang.
\newblock Generalized independent noise condition for estimating latent variable causal graphs.
\newblock {\em arXiv:2010.04917}, 2020.

\bibitem[\protect\citeauthoryear{Xie \bgroup \em et al.\egroup }{2024}]{xie2024generalized}
Feng Xie, Biwei Huang, Zhengming Chen, Ruichu Cai, Clark Glymour, Zhi Geng, and Kun Zhang.
\newblock Generalized independent noise condition for estimating causal structure with latent variables.
\newblock {\em Journal of Machine Learning Research}, 25:1--61, 2024.

\bibitem[\protect\citeauthoryear{Yang \bgroup \em et al.\egroup }{2024}]{yang2024qwen2}
An~Yang, Baosong Yang, Beichen Zhang, Binyuan Hui, Bo~Zheng, Bowen Yu, Chengyuan Li, Dayiheng Liu, Fei Huang, Haoran Wei, et~al.
\newblock Qwen2. 5 technical report.
\newblock {\em arXiv:2412.15115}, 2024.

\bibitem[\protect\citeauthoryear{Yao \bgroup \em et al.\egroup }{2021}]{yao2021survey}
Liuyi Yao, Zhixuan Chu, Sheng Li, Yaliang Li, Jing Gao, and Aidong Zhang.
\newblock A survey on causal inference.
\newblock {\em In TKDD}, 2021.

\bibitem[\protect\citeauthoryear{Ze{\v{c}}evi{\'c} \bgroup \em et al.\egroup }{2023}]{zevcevic2023causal}
Matej Ze{\v{c}}evi{\'c}, Moritz Willig, Devendra~Singh Dhami, and Kristian Kersting.
\newblock Causal parrots: Large language models may talk causality but are not causal.
\newblock {\em arXiv:2308.13067}, 2023.

\bibitem[\protect\citeauthoryear{Zeng \bgroup \em et al.\egroup }{2024}]{zeng2024survey}
Yan Zeng, Ruichu Cai, Fuchun Sun, Libo Huang, and Zhifeng Hao.
\newblock A survey on causal reinforcement learning.
\newblock {\em IEEE Transactions on Neural Networks and Learning Systems}, 2024.

\bibitem[\protect\citeauthoryear{Zhang \bgroup \em et al.\egroup }{2024a}]{zhang2024causal}
Congzhi Zhang, Linhai Zhang, Jialong Wu, Deyu Zhou, and Yulan He.
\newblock Causal prompting: Debiasing large language model prompting based on front-door adjustment.
\newblock {\em arXiv:2403.02738}, 2024.

\bibitem[\protect\citeauthoryear{Zhang \bgroup \em et al.\egroup }{2024b}]{zhang2024causaldistillationalleviatingperformance}
Shengyu Zhang, Ziqi Jiang, Jiangchao Yao, Fuli Feng, Kun Kuang, Zhou Zhao, Shuo Li, Hongxia Yang, Tat-Seng Chua, and Fei Wu.
\newblock Causal distillation for alleviating performance heterogeneity in recommender systems, 2024.

\bibitem[\protect\citeauthoryear{Zheng \bgroup \em et al.\egroup }{2024}]{processbench}
Chujie Zheng, Zhenru Zhang, Beichen Zhang, Runji Lin, Keming Lu, Bowen Yu, Dayiheng Liu, Jingren Zhou, and Junyang Lin.
\newblock Processbench: Identifying process errors in mathematical reasoning.
\newblock {\em arXiv:2412.06559}, 2024.

\bibitem[\protect\citeauthoryear{Zuo \bgroup \em et al.\egroup }{2024}]{zuo2024interventional}
Aoqi Zuo, Yiqing Li, Susan Wei, and Mingming Gong.
\newblock Interventional fairness on partially known causal graphs: A constrained optimization approach.
\newblock {\em arXiv:2401.10632}, 2024.

\end{thebibliography}
\newpage
\appendix

\section{Appendix}

\subsection{Datasets}
\label{Data}
We select the PROCESSBENCH (PB) dataset \cite{processbench}. It consists of 3,400 tests cases, primarily focused on competition and Olympiad-level math problems. Each test case contains a step-by-step solution with error location annotated by human experts, and half of the data is causalized, while the other half contains causal errors.

Specifically, it contains sample questions from the following four datasets:

\textbf{GSM8K} \cite{cobbe2021training} contains high quality linguistically diverse grade school math word problems.

\textbf{Math} \cite{hendrycks2021measuring} is a challenging competition math problems dataset. Each problem requires a complete step-by-step solution to arrive at the correct answer.

\textbf{OlympiadBench} \cite{he2024olympiadbench} is an Olympiad-level bilingual multimodal science benchmark that contains Olympiad-level math and physics competition problems, including the Chinese college entrance examination. Each problem requires expert-level annotations to complete step-by-step reasoning. We focus OlympiadBench's physics part in our experiment.

\textbf{Omni-MATH} \cite{gao2024omni} is a mathematics-focused, comprehensive and challenging benchmark specifically designed to assess LLMs' mathematical reasoning ability at the Olympiad level. It is rigorously manually annotated. The questions are carefully divided into more than 33 sub-areas covering more than 10 different difficulty levels.

\subsection{LLMs CoT reasoning: Causal view analysis}
Combining the results in Table~\ref{table1}, Figure~\ref{Qwen} and Figure~\ref{Deepseek}, we make the following analyses:

1.More complex problems require causal reasoning: As previously mentioned, the causal relationship between PB and correct answers is stronger than that of CoT in relatively complex datasets. A similar phenomenon is observed in Table~\ref{table1}. It indicates that problems requiring more complex reasoning demand reasoning paths with correct causal relationships for accurate answers. The greater improvements achieved by CauCoT over CoT on complex datasets compared to simpler datasets further confirm this conclusion.

2.Stronger language models have a greater ability to capture causal information: On the PB dataset, Qwen2.5-72b, and Deepseek-v3-37B achieves nearly 0.5 accuracy, indicating that the model can fully capture and understand half of the correct causal information in the dataset. This demonstrates from the perspective of LLMs capability that causality is a necessary factor in determining whether the reasoning is correct.
\subsection{ $\sigma$ evaluation}
\label{A2}
we analyze the numerical settings of the $\sigma$. For the implementation of the causal base model, we consistently choose Qianwen2.5-72b.

\begin{table}[H]
  \caption{$\sigma$ Evaluation. The first column lists the datasets used for evaluation. The second column shows the values of $\sigma$ set in the experiments, and the last column represents the proportion of CoT that successfully causalized.}
  \label{casualvalue}
  \centering
  \begin{tabular}{p{2cm}|p{2cm}|p{3cm}}
  \toprule
 \textbf{Dataset} & \textbf{$\sigma$ values}  &\textbf{Percents of successfully causalizing}\\
    \midrule
  GSM8K& 50   &  100\%  \\
   & 75  &  100\% 	 \\
   & 100    &   100\%   \\
  \midrule
  Math & 50  &   100\%  \\
   & 75   &  100\%   \\
   & 100    &  96\%   \\
  \midrule
  Olympiad & 50   &  100\% \\
   & 75   &  65\%   \\
   & 100    &   54\%    \\
  \midrule
  Omnimath & 50   & 100\% \\
   &  75 & 76 \%  \\
   & 100  & 69 \%\\

    \bottomrule
  \end{tabular}
\end{table}

As shown in the Table~\ref{casualvalue}, when the dataset is relatively complex, a higher $\sigma$ value makes updates more difficult to complete. This also highlights the necessity of setting $\sigma$ appropriately, allowing CauCoT to adjust the settings according to different scenarios to ensure feasibility.

\subsection{Ablation experiment}

We design baselines as :

\textbf{CauCoT-WOutRD}:
Removing $\gamma_{td}$, skip the step of validating the causal relationship between $Q$ and $C$.

\textbf{CauCoT-WOutRF}:
Removing Refine process, do not apply exhaustive thinking.

\textbf{CauCoT-WutLoop}:
Removing loop, trust the result of a single causalize attempt. 

Due to the minimal performance difference of CauCoT on GSM8K, we do not consider its performance on this dataset in the ablation experiments.

\label{A4}
\begin{figure}[H]
    \centering
    \includegraphics[width=1\linewidth]{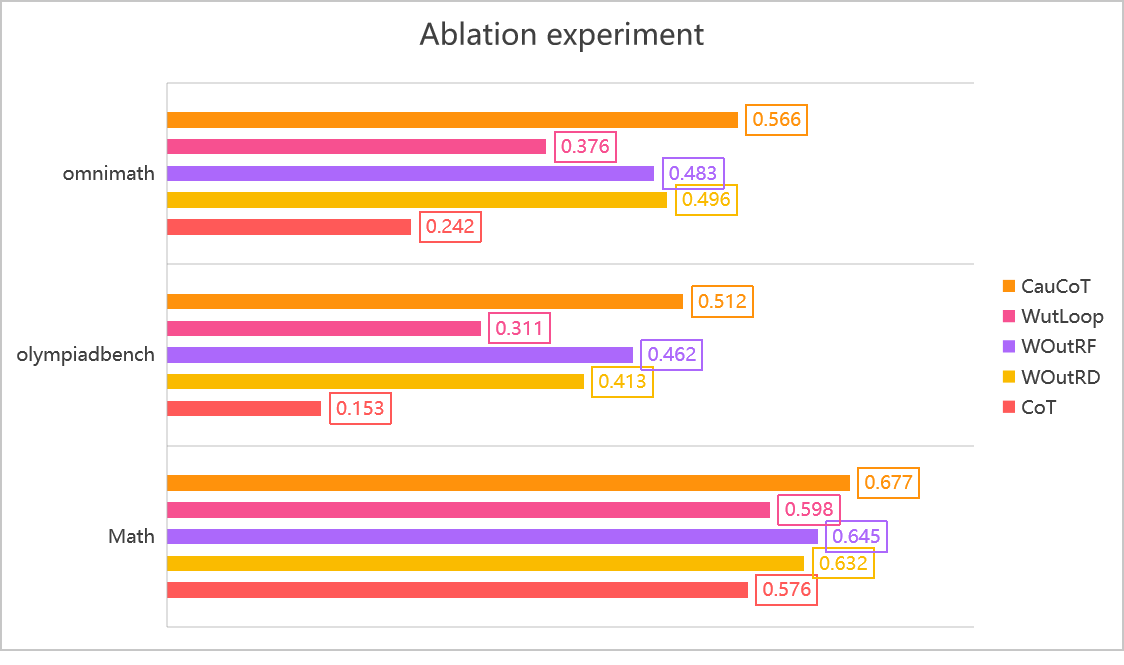}
    \caption{Ablation experiment}
    \label{ae}
\end{figure}

As shown in the Figure~\ref{ae}, the performance of the three baselines we proposed is inferior to that of CauCoT. The accuracy difference between WOutRD and CauCoT is equal to the percentage of errors in the first step of PB. This demonstrates that when there is an incorrect causal relationship between $Q$ and $C$, subsequent causalize steps are also difficult to complete. This confirms the necessity of setting the equation~\ref{8}. The performance of WOutRF proves the importance of slow thinking, which becomes even more apparent when dealing with relatively complex datasets (Omnimath and Olypiadbench). The performance of CauCoT-WutLoop demonstrates that existing LLMs still have an insufficient understanding of causality.

\subsection{Examples of causal errors in CoT}
\label{Eerror}

\textbf{Causality measure error \cite{Papana_2011,pearl2012measurement,chwialkowski2014wild,scheines2017measurement}.}

Causality measure error refers to the inaccuracy in causal inference caused by the incorrect measurement or transmission of causal variables within the CoT. For instance, as shown in the first example in Table \ref{Cerror}, this is a typical mathematical calculation problem. When an error occurs in calculation steps, incorrect causal variable data will be passed on to the next step, ultimately leading to an incorrect answer to the question.

\textbf{Collider error. \cite{burns2004collider,schneider2020collider,holmberg2022collider}.}

Collider error is an error caused by the inability to accurately measure the impact of two variables CoT on answering the question (the next step is the collision point). Like the second example in Table \ref{Cerror}, it is a problem that combines calculation and logical deduction. When both causal variables—division calculation and logical deduction based on the remainder—appear simultaneously, the next CoT step fails to accurately capture information from both, leading to an error in the calculation.

\textbf{Causal sensitivity error \cite{imai2010identification,diaz2013sensitivity,cinelli2019sensitivity}.}

Causal sensitivity error refers to the failure to accurately measure the impact of two causal variables in CoT on answering the question. It occurs when the next step fails to account for the sensitivity of the variables’ relationship, leading to an incorrect conclusion. For example, in the second case in Table \ref{Cerror}, it is a problem that involves both calculation and logical deduction. When both causal variables—such as division calculation and logical deduction based on the remainder—appear simultaneously, the next CoT step fails to accurately capture the influence of both, resulting in an error in the calculation.

\textbf{Mediation error \cite{pearl2014interpretation}.}

Mediation error occurs when CoT fails to correctly identify the mediator variable, which is essential in explaining the causal relationship between other variables. This error arises when CoT ignores the role of the mediator, leading to a misunderstanding of the causal pathway. For example, in the fourth case in Table \ref{Cerror}, CoT overlooks the iron pickaxe as a mediating variable and mistakenly establishes a direct causal relationship between two variables, resulting in an incorrect conclusion.

Moreover, we have emphasized the necessity of verifying the causal relationship between the first step and the $Q$, as discussed in formula \eqref{8}. For example, in question 3 in Table \ref{Cerror}, if the causal relationship between the first step and $Q$ is incorrect, it negatively impacts subsequent steps, leading to a failure in the reasoning process. This issue will also be further explored in Appendix~\ref{A3}.

\begin{table*}
  \caption{The table below provides examples of common causal errors in CoT that were corrected by CauCoT. The first column indicates the dataset to which the example belongs. The second column contains the content of the corresponding question. The third column specifies the step where the causal error occurred. The fourth column details the specific content of the error in that step, with the exact errors highlighted in \textcolor{red}{red}. The fifth column presents the corrected content by CauCoT, with the corrected parts marked in \textcolor{green}{green}.}
  \label{Cerror}
  \centering
  \begin{tabular}{|p{1.5cm}|p{4cm}|p{1cm}|p{4.5cm}|p{4.5cm}|}
  \toprule
 Dataset & \textbf{Question}  & \textbf{Steps}  & \textbf{Setps without causality}&  \textbf{Causalized CoT} \\
    \midrule
  GSM8K&  A company sold 4000 gallons of milk in jars to Mr. Marcellus' store at the cost of \$3.5 per gallon. However, Mr. Marcellus later realized 2/5 of the amount of milk he purchased had passed the expiry date and could not be sold. He returned the sour milk to the company and ordered a refund. Calculate how much he got in refunds.
   & Step 3 &"...\textcolor{red} {Total gallons purchased = 2/5 * 4000 = 800 gallons."},
"Now, we need to calculate the cost of the expired milk that needs to be refunded: Cost of expired milk = Cost per gallon * Amount of expired milk = \$3.5 $*$ 800 = \$2,800.",
"Therefore, Mr. Marcellus got a refund of \$2,800 for the unsellable milk. [The answer is \$2800."\textcolor{red}{×}]
   & "...\textcolor{green}{Total gallons purchased by the fraction of expired milk: 4000 gallons $*$ (2/5) = 1600 gallons.}",
 "Now, we need to calculate the cost of the expired milk that needs to be refunded: Cost of expired milk = Cost per gallon * Amount of expired milk = \$3.5 $*$ 1600 = \$5,600.",
"Therefore, Mr. Marcellus got a refund of \$5,600 for the unsellable milk. [The answer is \$5,600."\textcolor{green}{\checkmark}]
  \\
  \midrule
  Math& Carlos Montado was born on Saturday, November 9, 2002.  On what day of the week will Carlos be 706 days old? (Spell out the entire day of the week.
&Step 2&"...
 \textcolor{red}{Divide 706 by $7: \frac{706}{7} = 101$ with a remainder of 1. This means that 706 days from his birth date is equivalent to adding one day to his birth day.}",
 "Since Carlos was born on a Saturday, we add 1 day to Saturday: Saturday + 1 day = Sunday.",
 "Therefore, the day of the week Carlos will be 706 days old is: [Sunday."\textcolor{red}{×}]
	& "...\textcolor{green}{Since the result is 100 and reminder is 6, reduce 1 day to Saturday, the day Carlos was born.} This will give us the day of the week he will be 706 days old.",
 "As Carlos was born on a Saturday, we reduce 1 day to Saturday: Saturday - 1 day = Firday.",
 "Therefore, the day of the week Carlos will be 706 days old is: [Firday."\textcolor{green}{\checkmark}]
  \\
  \midrule
  Olympiad & Bobby, Peter, Greg, Cindy, Jan, and Marcia line up for ice cream. In an acceptable lineup, Greg is ahead of Peter, Peter is ahead of Bobby, Marcia is ahead of Jan, and Jan is ahead of Cindy. For example, the lineup with Greg in front, followed by Peter, Marcia, Jan, Cindy, and Bobby, in that order, is an acceptable lineup. Compute the number of acceptable lineups.",
 &  Step 1	 &  "To determine the number of acceptable lineups for Bobby, Peter, Greg, Cindy, Jan, and Marcia, we need to consider the constraints given: First, Greg is ahead of Peter. Second, Peter is ahead of Bobby. Third, Marcia is ahead of Jan. Fourth, Jan is ahead of Cindy. These constraints imply that the relative positions of \textcolor{red}{Greg, Peter, Bobby, Marcia, Jan, and Cindy must be in the following order: Greg, Peter, Bobby, Marcia, Jan, Cindy.}"..."Thus, the number of acceptable lineups is [35"\textcolor{red}{×}]) & "To determine the number of acceptable lineups for Bobby, Peter, Greg, Cindy, Jan, and Marcia, we need to consider the constraints given: First, Greg is ahead of Peter. Second, Peter is ahead of Bobby. Third, Marcia is ahead of Jan. Fourth, Jan is ahead of Cindy. These constraints imply that the relative positions of\textcolor{green}{ Greg, Peter, Bobby, Marcia, Jan, and Cindy must be in the following two order: Greg, Peter, Bobby and Marcia, Jan, Cindy.}"..." The number of ways to select 3 positions from 6 is given by the combination formula acceptable lineups is [20."\textcolor{green}{\checkmark}]
  \\  \midrule
  GSM8K & One dwarf can mine 12 pounds of ore per day with his bare hands. He can mine twice as much with an iron pickaxe and 50\% more with a steel pickaxe than with an iron pickaxe. How many pounds of ore can 40 dwarves with steel pickaxes mine in a month with 30 days? & Step 2&  "... \textcolor{red}{First, we calculate the amount of ore one dwarf can mine with a steel pickaxe.} We know that one dwarf can mine 50\% more with a steel pickaxe than with an iron pickaxe" ..." So, the total amount of ore mined by 40 dwarves with steel pickaxes in a month is:[21600."\textcolor{red}{×}] & "...\textcolor{red}{First, we calculate the amount of ore one dwarf can mine with a iron pickaxe.} We know that one dwarf can mine twice as much with an with an iron pickaxe..."So, the total amount of ore mined by 40 dwarves with steel pickaxes in a month is:[43,200. "\textcolor{green}{\checkmark}]"
  \\
    \bottomrule
  \end{tabular}
\end{table*}

\section{Technical report}

\subsection{Details of LLMs}

We updated the reasoning chains in the ProcessBench dataset (https://huggingface.co/datasets/Qwen/ProcessBench) using the Qwen2.5-72B model (https://huggingface.co/Qwen/Qwen2.5-72B) via the Transformer Library (https://huggingface.co/docs/transformers/en/index). The hyperparameters for generation remained consistent across all prompt methods. During the update process, we utilized a pipeline as a high-level helper, and set the task type to "text-generation" while keeping other hyperparameters at their default values.

Additionally, we compared the question-answering performance before and after updates using large models: Qwen2.5-3B, Qwen2.5-7B, Llama3-8B, Qwen2.5-72B, and Deepseek-v3-37B. For the Qwen2.5-3B, Qwen2.5-7B, Llama3-8B, and Qwen2.5-72B models, we set the voting count to 8, based on prior work \cite{processbench}. The final output was determined by majority voting. 

Link of LLMs intructions:

Qwen2.5-3B:

(https://huggingface.co/Qwen/Qwen2.5-3B).

Qwen2.5-7B:

(https://huggingface.co/Qwen/Qwen2.5-7B).

Llama3-8B:

(https://huggingface.co/meta-llama/Meta-Llama-3-8B).

Qwen2.5-72B:

(https://huggingface.co/Qwen/Qwen2.5-72B).

Deepseek-v3-37B:

(https://github.com/deepseek-ai/DeepSeek-V3).

More specifically, the generation hyperparameters remained consistent across all prompt methods, as shown in the table~\ref{HPLLM}:
\begin{table*}[htp]
\caption{Hyperparameter Settings for Different Models}
\label{HPLLM}
\centering
\begin{tabular}{|c|c|c|c|c|}
\hline
\textbf{Hyperparameter} & \textbf{Qwen2.5-3B} & \textbf{Qwen2.5-7B} & \textbf{Llama3-8B} & \textbf{Qwen2.5-72B} \\
\hline
temperature & 1 & 1 & 1 & 1 \\
\hline
top p & 0.9 & 0.9 & 0.9 & 0.9 \\
\hline
trust remote code & True & True & True & True \\
\hline
times of votes & 8 & 8 & 8 & 8 \\
\hline
max tokens & 32768 & 32768 & 32768 & 32768 \\
\hline
tensor parallel size & 4 & 4 & 4 & 8 \\
\hline
\end{tabular}

\end{table*}

Due to the large size of the Deepseek-v3-37B model, we use the OpenAI Python API library and an API key obtained from Deepseek's official request to call it. The generation hyperparameters are shown in the Table~\ref{HPDEEP}:

\begin{table}[H]
\caption{Hyperparameters of Deepseek-v3-37B in experiments}
\label{HPDEEP}
\centering
\begin{tabular}{|c|c|}
\hline
\textbf{Hyperparameter} & \textbf{Value}\\
\hline
stream & False \\
\hline
model & deepseek-chat\\
\hline
\end{tabular}

\end{table}

\end{document}